\documentclass{article}

\usepackage[preprint, nonatbib]{neurips_2019}

\usepackage[utf8]{inputenc} 
\usepackage[T1]{fontenc}    
\usepackage{hyperref}       
\usepackage{url}            
\usepackage{booktabs}       
\usepackage{amsfonts}       
\usepackage{nicefrac}       
\usepackage{microtype}      
\usepackage{algpseudocode}
\usepackage[algo2e,boxruled,linesnumbered]{algorithm2e}
\usepackage{algorithm}

\usepackage{multirow}
\usepackage{float}
\usepackage{graphicx}
\usepackage{mathtools}
\usepackage{wrapfig}
\usepackage[font=footnotesize]{caption}

\title{Adaptive Attention Span in Computer Vision}

\author{%
  Jerrod Parker\thanks{Equal contribution} \hspace{1cm}Shakti Kumar\footnotemark[1]\hspace{1cm}Joe Roussy\footnotemark[1]\\
  Department of Computer Science \\
  University of Toronto \\
  \texttt{\{jparker,shaktik,jroussy\}@cs.toronto.edu} \\
}

\begin{document}

\maketitle

\begin{abstract}
Recent developments in Transformers for language modeling have opened new areas of research in computer vision. Results from late 2019 showed vast performance increases in both object detection and recognition when convolutions are replaced by local self-attention kernels. Models using local self-attention kernels were also shown to have less parameters and FLOPS compared to equivalent architectures that only use convolutions. In this work we propose a novel method for learning the local self-attention kernel size. We then compare its performance to fixed-size local attention and convolution kernels. The code for all our experiments and models is available at \url{https://github.com/JoeRoussy/adaptive-attention-in-cv}.

\end{abstract}

\section{Introduction}

Transformers have been shown to successfully model natural language through the use of self-attention to capture long-range dependencies in text \cite{Transformer}. It is natural to consider the extension of Transformers in computer vision tasks which share a great deal with language modelling such as high correlation between adjoining pixels and the need to capture long-range pixel dependencies. Recent work has shown that replacing convolutions with attention kernels increases the performance of vision models while using considerably less parameters and FLOPS \cite{StandAlone}. Additionally, recent research \cite{AdaptiveAttention} has shown that learning an adaptive attention span in Transformers improves performance in language modeling by allowing the model to attend over long ranges without increasing computation. Thus, we hypothesise that learning an adaptive attention span should improve computer vision models by capturing long-range dependencies between pixels more easily than fixed-size attention kernels.

The rest of the paper is organized as follows. In Section 2 we discuss the background and some seminal work in using attention within computer vision models. Section 3 describes our proposed adaptive span model. In section 4, we conduct extensive experiments to compare our adaptive attention method to the non-adaptive attention of \cite{StandAlone} in addition to pure convolutions for image classification on CIFAR100 \cite{cifar100}. We compare how the performance of these kernel primitives scale in terms of parameters, FLOPS, and training samples. This is followed by discussion of our findings as well as limitations and future improvements.

\section{Related Work}
Self-attention architectures have become widely used in natural language applications due to their ability to capture long-range dependencies between words. Recently, self-attention has been successfully applied to computer vision tasks. \cite{AACNN} augmented convolution layers with attention layers to provide a more global representation of the inputs. This method had high computational cost as the time complexity of self-attention grows quadratically with the input size which is very large for images. A lower complexity approach was introduced by \cite{StandAlone} which replaces each convolution kernel with an attention kernel of the same size. They found that using attention kernels in the bottleneck blocks of a ResNet \cite{ResNet} allowed their model to outperform a convolution-only version of the same architecture.

Within language modeling, \cite{AdaptiveAttention} improved feasibility of using attention on very long character sequences by allowing each attention head within a self-attention layer to learn the context size that it should attend to. This prevents each attention head from having to attend to every other element which results in faster training. As a result, the model matches the performance of the previous state of the art \cite{TXL} with a much lower parameter count.

In this work, our main contribution is applying adaptive attention span to computer vision. Our hypothesis is that we can keep some of the benefits of attention augmented CNNs \cite{AACNN} while maintaining computational complexity similar to local self-attention kernels \cite{StandAlone} by learning the minimum necessary span of each attention head.

\begin{figure}
    \centering
    \includegraphics[scale=0.6]{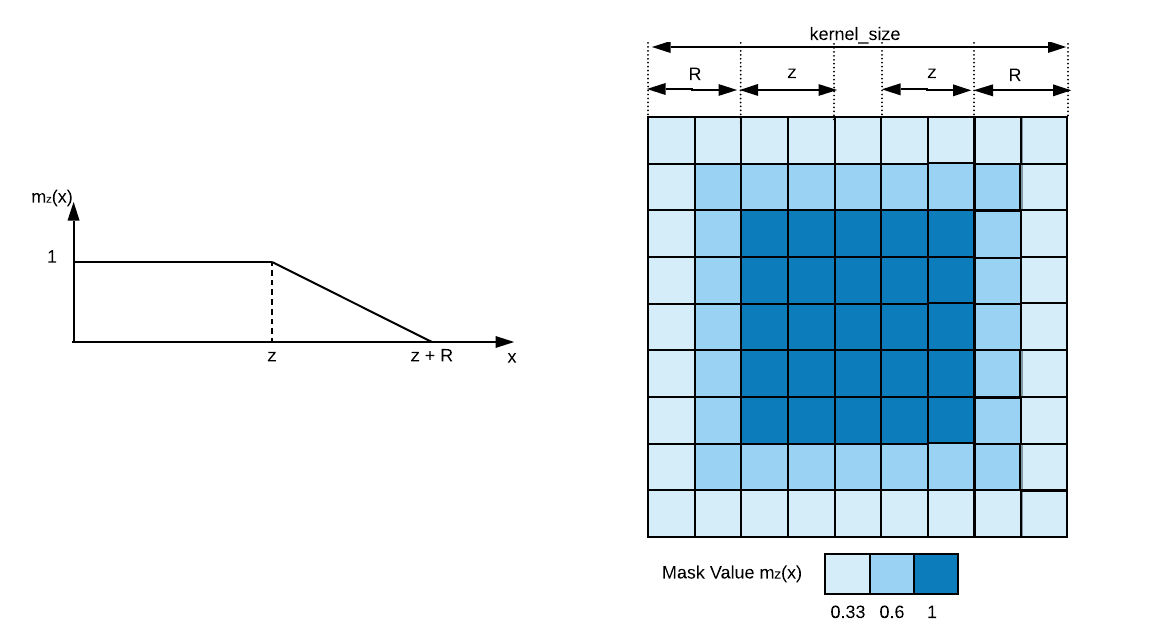}
    \caption{\textbf{Left:} The 1D attention mask used in \cite{AdaptiveAttention} shows the mask $m_z$ as a function of span $z$. \textbf{Right:} Our generalization of the 1D mask to 2D. This is an example of the attention mask when $z=2$, $R=2$.}
    \label{fig:mask_function}
\end{figure}{}

\section{Method}
In our proposed method, we first swap convolutions with local attention kernels as done in \cite{StandAlone} and then incorporate a learnable kernel size, as shown on line 9 of Algorithm \ref{alg:algorithm1} and Figure \ref{fig:mask_function}. Local attention works similarly to convolution in that the value at a pixel is computed as a function of the pixels surrounding it. We compute a query projection for the pixel at the center of the kernel and key projections for all the pixels in the kernel. The query then attends to all the keys to obtain attention logits. Based on these results, we return a weighted combination of the value projections for each pixel in the kernel. A diagram explaining local attention is shown in Figure \ref{fig:stand_alone_figure3} of Appendix A.1.

For each pixel $x_{ij}$ we compute the adaptive attention result $y_{ij}$ for a single attention head as:
\begin{equation}
    y_{ij} = \sum_{(r,s) \in N_{k(i,j)}} \frac{e^{a_{rs}}M_{rs}}{\sum_{(c,d) \in N_{k(i,j)}} e^{a_{cd}}M_{cd}} v_{rs}
    \label{eqn:firsteqn}
\end{equation}{}

where $a_{rs} = q_{ij}^T k_{rs}$, $q_{ij} = Qx_{ij}$, $k_{ij} = Kx_{ij}, v_{ij}=Vx_{ij}$, and $N_{k(i,j)}$ is the set of pixels in a square region of length $kernel\_size$ centered at $(i, j)$. $M_{rs}$ is the attention mask for this head at index $(r, s)$ of the kernel and depends on the ramp size $R$ and attention span $z$ as shown in Figure \ref{fig:mask_function}. In practice we use multiple heads to capture different representations of the input. The attention logits are masked separately for each head so the kernel size that we need to compute for a layer depends on the maximum span over the heads for that layer.

\begin{figure}
    \centering
    \includegraphics[scale=0.64]{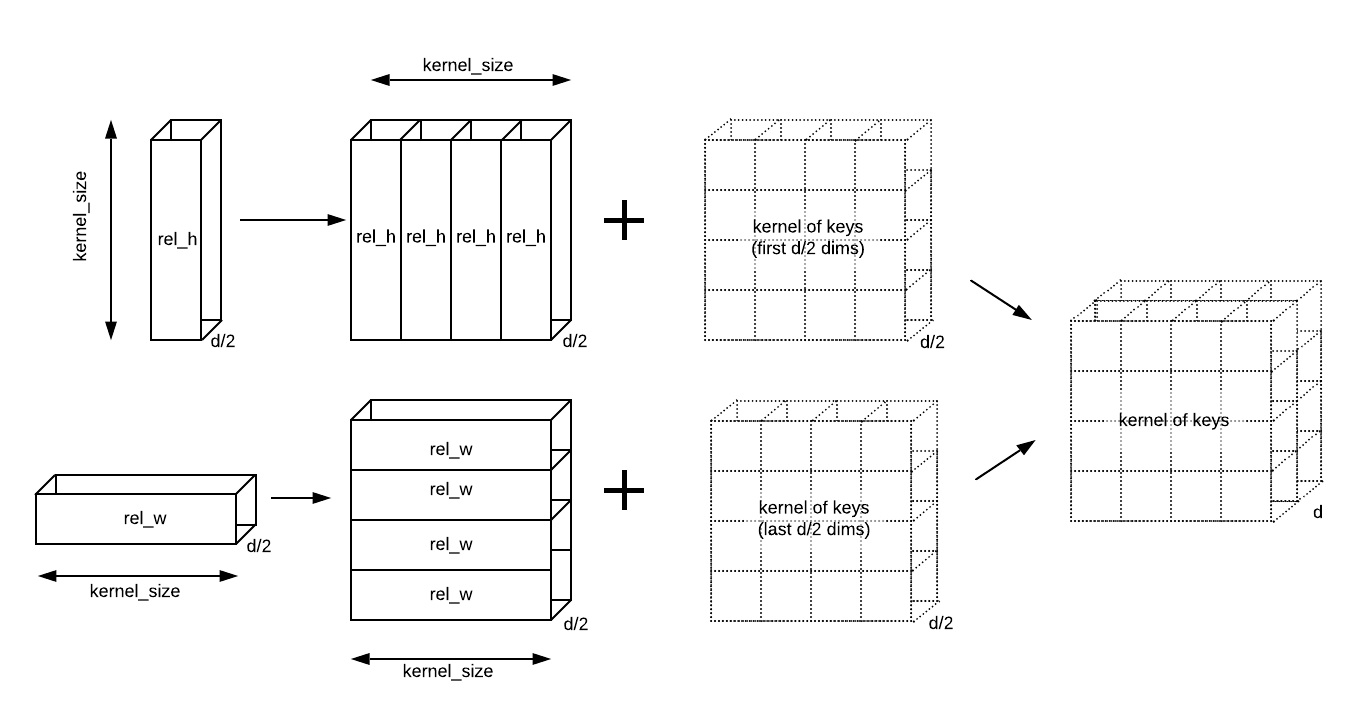}
    \caption{Relative width embeddings are applied to the first $\frac{d}{2}$ dimensions of the key vectors and relative height embeddings are applied to the last $\frac{d}{2}$ dimensions. These are applied to the key vector corresponding to a given pixel based on the location of the pixel in the kernel. Note that $rel\_w$ and $rel\_h$ shown on the left side of image are the middle $kernel\_size$ elements of the width and height positional embedding matrices which allow the dynamic kernel sizes.}
    \label{fig:relative_emb}
\end{figure}

Our relative positional embedding is very similar to that of \cite{StandAlone}. Since the size of the attention kernels depend on the maximum attention span, both the height and width embedding vectors have a size of $input\_size$. These embeddings are relative to the center pixel of the kernel $x_{i,j}$ as shown in Figure \ref{fig:relative_emb}. This allows each element within the kernel to know its location relative to other pixels in the kernel. To add these positional embeddings to a kernel, we take the middle $kernel\_size$ elements of those two vectors and then add them to the key projections as is done in \cite{StandAlone} and shown in Figure \ref{fig:relative_emb}. Algorithm \ref{alg:algorithm1} explains the detailed steps for our approach.

{}

\newfloat{algorithm}{t}{lop}
\begin{algorithm}
 \SetAlgoLined
 \begin{tabbing}
\nl Input: \=pixel location$(i, j)$,\\
\nl \> $Q, K, V \in {\rm I\!R}^{d \times d}$ query, key and value matrices  \\

\nl \> $X \in {\rm I\!R}^{input\_size \times input\_size \times d}$ the input \\
\nl \> $H, W \in {\rm I\!R}^{input\_size \times \frac{d}{2}}$ are height and width embedding matrices respectively \\
\nl \> $z$, the learnable attention span shown in Figure \ref{fig:mask_function}\\
\nl \> $R > 0$ a hyperparameter denoting ramp length\\

\end{tabbing}
 // Calculate the attention span \\
\nl $max\_size = z + R$ \\
\nl $max\_size = max(0, min(max\_size, input\_size))$ \\
\nl $kernel\_size = 2*max\_size + 1$ \\
\nl $mask = CreateAdaptiveMask(kernel\_size)$ // as done in Figure \ref{fig:mask_function} \\
\nl $N_{k(i,j)}=$ the set of pixels in a square region of length $kernel\_size$ centered at $(i, j)$ \\
\nl $start\_ind = (d - kernel\_size)/2$ \\
\\
 // Get the middle $kernel\_size$ rows of $H$ and columns of $W$ \\
\nl $rel\_h = H[start\_ind: d-start\_ind, :], rel\_w = W[start\_ind: d-start\_ind, :]$ \\
\nl Add relative positional encoding to the keys as shown in Figure \ref{fig:relative_emb} \\
\nl Compute output y as done in Equation \ref{eqn:firsteqn} // Compute adaptive attention  \\
\nl \Return{y}

 \caption{Adaptive Attention for 1 Head for a Given Pixel}
 \label{alg:algorithm1}
\end{algorithm}

\section{Experiments}
We compare the relative performances of our proposed adaptive attention span kernel, non-adaptive attention kernels \cite{StandAlone} and convolution kernels on CIFAR100 \cite{cifar100} using the ResNet architecture \cite{ResNet}. In the models that use attention, we swap out the convolution kernels with fixed-span or adaptive span kernels as, described in section 3.

The hyperparameters for each model are chosen based on their validation set accuracy after 100 epochs of training and the test set accuracies are reported. All models are trained using stochastic gradient descent \cite{SGD} with 0.9 Nesterov momentum \cite{nesterov}. The learning rates were decayed using cosine annealing after an initial warmup. We use 3 different architectures---small, medium and large---with a varying number of layers and channels as described in Appendix A.2. The hyperparameters used for all our experiments can be found in Appendix A.3.

From the plots of Figure \ref{fig:accuracyPlot}, we can see that the CNN scales better with parameters, FLOPS, and training data than the attention-based kernels. Surprisingly, we see that using adaptive kernel sizes causes the model to scale worse than non-adaptive kernels in terms of model size and number of training samples.

For the medium adaptive model with 4 bottleneck layers, the maximum kernel sizes learned were 7, 7, 5, and 5 respectively. On the other hand, the non-adaptive attention model performed best with a kernel size of 5 at each layer. We see that the adaptive model learns larger kernel sizes in earlier layers where the input sizes are largest. However, this does not translate into increased performance over non-adaptive attention kernels.

\begin{figure}
    \centering
    \includegraphics[scale=0.3]{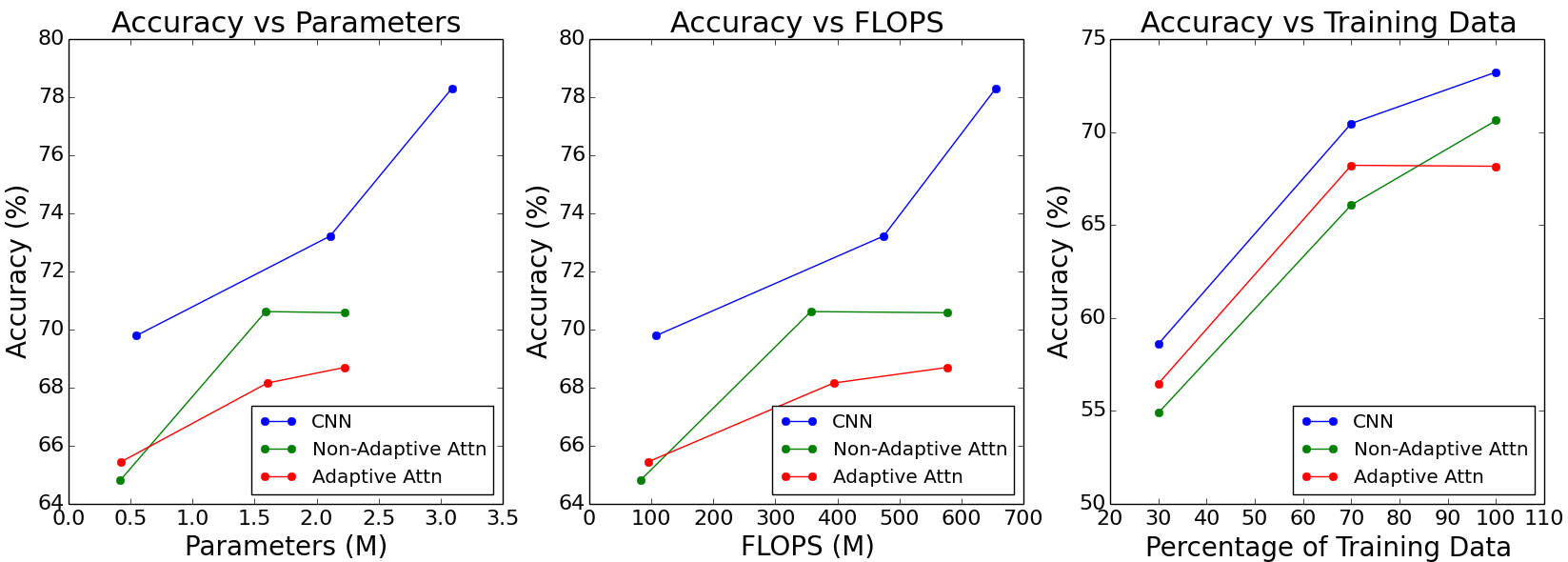}
    \caption{
    \textbf{Left and Center}: Comparing relative performance of small, medium, and large models with different kernels. \textbf{Right}: Comparing performance of the medium-sized models with the proportion of training data used.
    }
    \label{fig:accuracyPlot}
\end{figure}{}

\begin{table}[ht]
    \scriptsize
  \centering
    \captionsetup{justification=centering, skip=2pt}
  \begin{tabular}{|l|l|l|l|l|l|l|l|l|l|}

    \hline
    \multirow{3}{*}{} &
      \multicolumn{3}{c|}{\textbf{Small}} &
      \multicolumn{3}{c|}{\textbf{Medium}} &
      \multicolumn{3}{c|}{\textbf{Large}} \\
    & \textbf{FLOPS} & \textbf{Params} & \textbf{Acc.} & \textbf{FLOPS} & \textbf{Params} & \textbf{Acc.} & \textbf{FLOPS} & \textbf{Params} & \textbf{Acc.}\\

    & \textbf{(M)} & \textbf{(M)} & \textbf{(\%)} & \textbf{(M)} & \textbf{(M)} & \textbf{(\%)} & \textbf{(M)} & \textbf{(M)} & \textbf{(\%)}\\

    \hline
    Conv & 107 & 0.54 & 69.8 & 474 & 2.10 & 73.2 & 655 & 3.09 & 78.3 \\
    \hline
    Non-Adaptive & 82.7 & 0.42 & 64.8 & 357 & 1.59 & 70.6 & 499 & 2.23 & 70.6 \\
    \hline
    Adaptive & 95.0 & 0.42 & 65.4 & 394 & 1.60 & 68.2 & 578 & 2.26 & 68.7 \\
    \hline
  \end{tabular}
  \caption{Results of using convolution, non-adaptive attention, and adaptive attention kernels on CIFAR-100 where Acc. is the test set classification accuracy.}
  \label{tab:return_table}
  \par
\end{table}

\section{Discussion}

The wall clock time for training our adaptive attention model is much larger than the training time for the same architecture using convolutions. For example, our medium-sized adaptive model takes 8x longer to train per epoch than the fully convolutional model. We think this is primarily due to a lack of hardware support for attention kernels. The lack of scalability of our implementation prevented us from running our method on the same networks used in \cite{StandAlone}, which are an order of magnitude larger in terms of parameters and showed superior performance of local attention over convolution. However, we see that convolutions outperform attention kernels for smaller models with less than 3M parameters. Hence, the relative performance of adaptive and non-adaptive kernels in the larger models of \cite{StandAlone} is unknown.

Furthermore, computational limitations prevented us from performing an exhaustive hyperparameter search for the adaptive span model, as was done for the convolutional and non-adaptive models. As such, we may be understating the performance of the adaptive attention span model compared to the other models in our experiments.

Additionally, the ResNet architecture may not be optimal for attention-based kernels. For a given computational budget, attention allows a larger kernel size than convolution. This allows the model to capture long-range dependencies in a smaller number of layers. Hence we expect that the early downsampling in ResNet may not be necessary in order to capture global information. As such, future works can improve our results by finding an optimal architecture for all-attention kernels by using a neural-architecture search as was proposed in \cite{StandAlone}.

Our experiments have been limited to image classification using CIFAR100. Future work should consider testing on ImageNet and COCO object detection datasets to explore the performance of adaptive attention span models on more difficult tasks.


\section{Conclusion}

In this paper, we present a novel 2D attention kernel which learns an optimal span for each head. We compare the relative performance of convolutions, non-adaptive attention, and adaptive attention kernels on CIFAR100. We found that attention kernels do not outperform convolutions on small models with less than 3M parameters. Furthermore, for models of this scale, learning an adaptive span does not provide any benefits over fixed-span attention kernels. We look forward to exploring the use of adaptive attention span in computer vision on larger models in the future.

\bibliographystyle{unsrt}
\bibliography{dissertationbib}
\newpage
\appendix
\section{Appendix}
\subsection{Local Self-Attention}
\begin{figure}[H]
    \centering
    \includegraphics[scale=0.65]{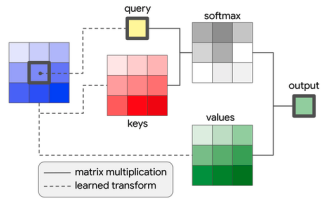}
    \caption{Figure taken from \cite{StandAlone} showing how local attention is computed for a kernel size of 3.}
    \label{fig:stand_alone_figure3}
\end{figure}{}

\subsection{Architecture Details}
Each model uses the same convolution stem which consists of 32 channels with kernel size 3x3 and a stride of 1. The number of channels for each of the succeeding residual layers of each model size are:

\textbf{Small:} 3 layers with 32, 64, and 128 channels respectively.\\
\textbf{Medium:} 4 layers containing 32, 64, 128, and 256 channels respectively.\\
\textbf{Large:} 9 layers containing 32, 64, 64, 64, 128, 128, 128, 128, 256 channels respectively.

The layers above are then followed by a linear output layer and then by a softmax activation to obtain probabilities for each class.

\subsection{Additional Training Details}
For all our experiments, we used Stochastic gradient descent with Nesterov momentum 0.9 and decayed the learning rate via cosine annealing after an initial warmup of 10 epochs. The models using convolution had a learning rate of 0.2 and weight decay 0.0001 while the attention-based models used a learning rate of 0.05 and weight decay 0.0005. Each model was trained on 2 NVIDIA P100 GPUs for 100 epochs using a batch size of 50. For the attention kernels, we used 4 heads because we found that using 8 heads resulted in similar validation accuracies, despite requiring significantly more wall clock time for training. The kernel size used in the non-adaptive attention models were 5x5 and the convolution kernels were 3x3. All adaptive attention models used a ramp size of 2 and initial kernel sizes of 4x4.

\end{document}